\documentclass[10pt,twocolumn,letterpaper]{article}

\usepackage{cvpr}
\usepackage{times}
\usepackage{epsfig}
\usepackage{graphicx}
\usepackage{amsmath}
\usepackage{amssymb}

\usepackage{multirow}

\usepackage[breaklinks=true,bookmarks=false,colorlinks]{hyperref}
\usepackage{cleveref}

\cvprfinalcopy 

\def\cvprPaperID{7753} 

\ifcvprfinal\fi
\begin{document}

\title{Multi-Scale Progressive Fusion Network for Single Image Deraining}

\author{Kui Jiang$^1$~~~~Zhongyuan Wang$^1$\thanks{Corresponding author}~~~~Peng Yi$^1$~~~~Chen Chen$^2$$^*$\\
Baojin Huang$^1$~~~~Yimin Luo$^3$~~~~Jiayi Ma$^1$~~~~Junjun Jiang$^4$\\
$^1$Wuhan University~~~~$^2$University of North Carolina at Charlotte\\
$^3$King's College London~~~~$^4$Harbin Institute of Technology\\
}
\maketitle
\thispagestyle{empty}

\begin{abstract}
Rain streaks in the air appear in various blurring degrees and resolutions due to different distances from their positions to the camera. Similar rain patterns are visible in a rain image as well as its multi-scale (or multi-resolution) versions, which makes it possible to exploit such complementary information for rain streak representation. In this work, we explore the multi-scale collaborative representation for rain streaks from the perspective of input image scales and hierarchical deep features in a unified framework, termed multi-scale progressive fusion network (MSPFN) for single image rain streak removal. For similar rain streaks at different positions, we employ recurrent calculation to capture the global texture, thus allowing to explore the complementary and redundant information at the spatial dimension to characterize target rain streaks. Besides, we construct multi-scale pyramid structure, and further introduce the attention mechanism to guide the fine fusion of this correlated information from different scales. This multi-scale progressive fusion strategy not only promotes the cooperative representation, but also boosts the end-to-end training. Our proposed method is extensively evaluated on several benchmark datasets and achieves state-of-the-art results. Moreover, we conduct experiments on joint deraining, detection, and segmentation tasks, and inspire a new research direction of vision task-driven image deraining. The source code is available at \url{https://github.com/kuihua/MSPFN}.

\end{abstract}
\section{Introduction}
Due to substantial degradation of the image content in rain images and videos, traditional image enhancement algorithms~\cite{teichmann2018multinet} struggle to make desirable improvements on image quality. Therefore, developing specialized solutions for image deraining is imperative to a wide range of tasks~\cite{He2017ICCV}, \eg object detection and semantic segmentation.
\begin{figure}[t]
\centering
\includegraphics[width=2.6in]{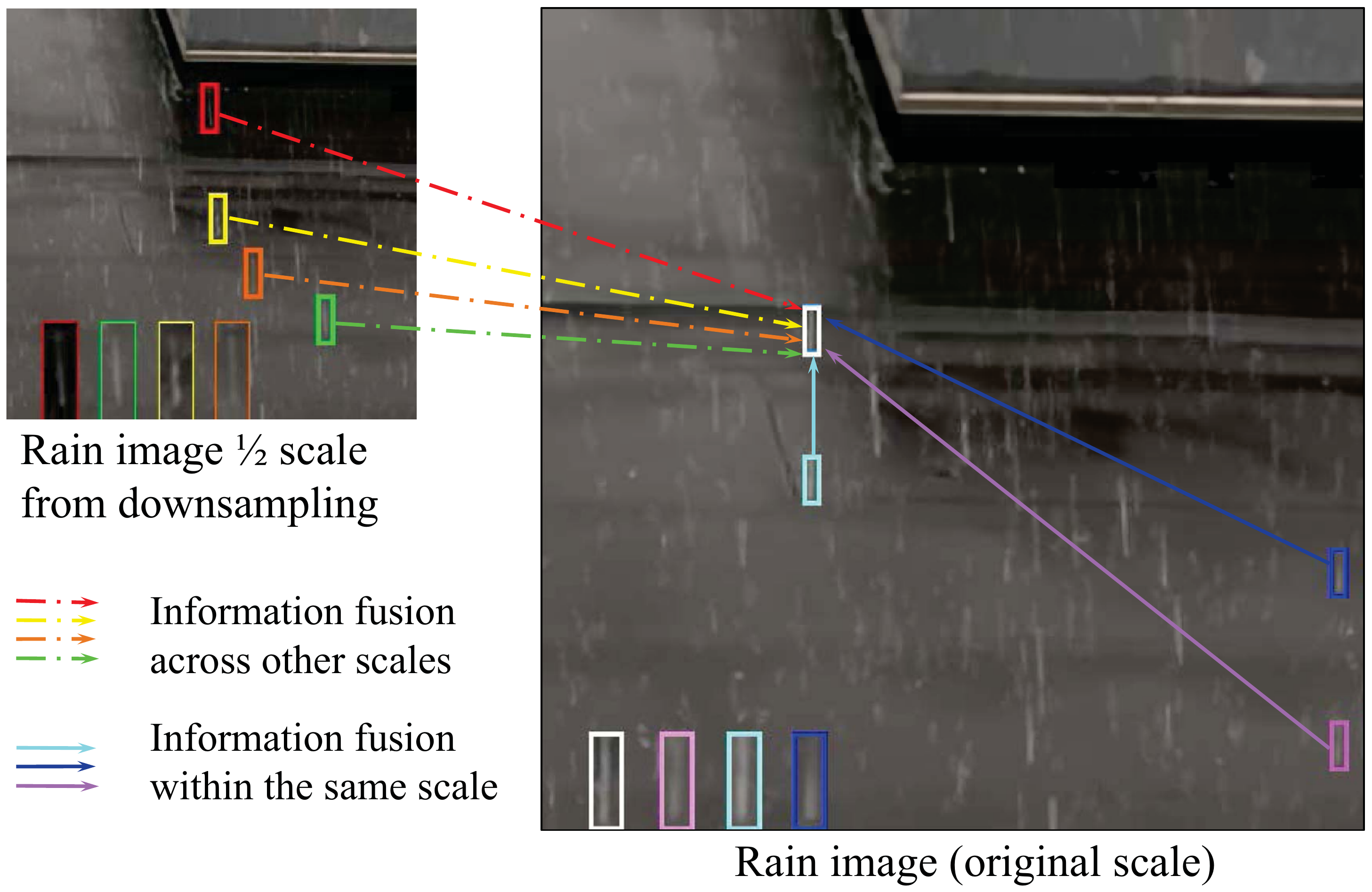}
\caption{Demonstration of the collaborative representation of rain streaks.
Specifically, similar rain patterns among rain streaks, both within the same scale (highlighted in cyan, pink and dark blue boxes) or cross different scales (highlighted in red, yellow, orange and green boxes), can help reconstruct the target rain streak (white box in the original rain image) with the complementary information (\eg similar appearance, formation, etc.).}
\label{fig:example}
\end{figure}

Traditional deraining methods~\cite{bossu2011rain, barnum2010analysis, chen2013generalized, garg2005does, xu2012improved} use simple linear-mapping transformations and are not robust to variations of the input~\cite{hanocka2019meshcnn}, \eg, rain streaks with various directions, densities and sizes. Recently, deep-learning based methods~\cite{fu2017clearing, yang2017deep,li2017single} which operate with convolutional and non-linear layers have witnessed remarkable advantages over traditional methods. Despite obvious improvements on feature representation brought by those methods~\cite{fu2017clearing, li2017single}, their single-scale frameworks can hardly capture the inherent correlations of rain streaks across scales.

The repetitive samples of rain streaks in a rain image as well as its multi-scale versions (multi-scale pyramid images) may carry complementary information (\eg similar appearance) to characterize target rain streaks. As illustrated in Fig.~\ref{fig:example}, the rain streaks (highlighted in the white box) in the original rain image share the similar rain patterns with the rain streaks (highlighted in the cyan, pink and dark blue boxes) at different positions as well as those (highlighted in the red, yellow, orange and green boxes) in the 1/2 scale rain image. Therefore, rain streaks both from the same scale (solid arrows) and across different scales (dashed arrows) encode complementary or redundant information for feature representation, which would help deraining in the original image. This correlation of image contents across scales has been successfully applied to other computer vision tasks~\cite{glasner2009super, yang2010exploiting}. Recently, authors in~\cite{fu2019lightweight,zhengresidual} construct pyramid frameworks to exploit the multi-scale knowledge for deraining. Unfortunately, those exploitations fail to make full use of the correlations of multi-scale rain streaks (although restricted to a fixed scale-factor of 2~\cite{glasner2009super}). For example, Fu~\etal~\cite{fu2019lightweight} decompose the rain image into different pyramid levels based on its resolution, and then individually solve the restoration sub-problems at the specific scale space through several parallel sub-networks. Such decomposition strategy is the basic idea of many recurrent deraining frameworks~\cite{li2018recurrent}. Unlike~\cite{fu2019lightweight} completing the deraining task from each individual resolution level, Zheng~\etal~\cite{zhengresidual} present a density-specific optimization for rain streak removal in a coarse-to-fine fashion, and gradually produce the rain-free image stage-by-stage~\cite{lai2017deep}. However, there are no direct communications of the inter-level features across cascaded pyramid layers except for the final outputs, thus failing to take all-rounded advantages of the correlated information of rain streaks across different scales. Consequently, these methods~\cite{fu2019lightweight, zhengresidual} are still far from producing the desirable deraining results with the limited exploitation and utilization of multi-scale rain information.

\begin{figure*}[!htpb]
\centering
\includegraphics[width=6.2in]{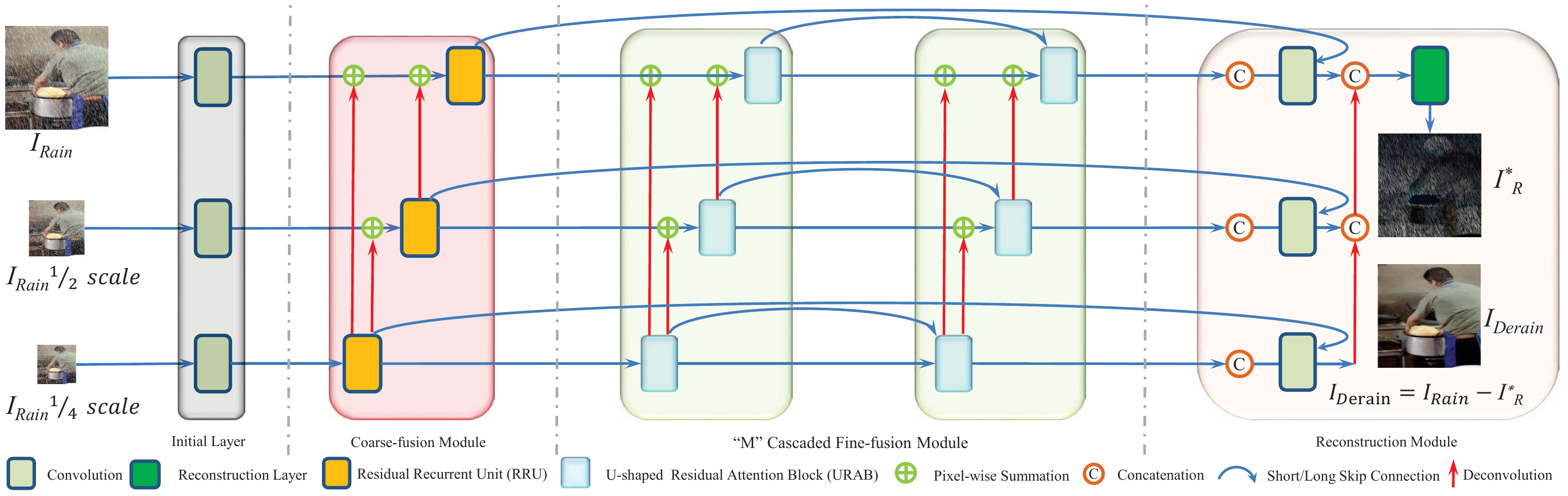}
\caption{Outline of the proposed multi-scale progressive fusion network (MSPFN). We set the pyramid level to 3 as an example. MSPFN consists of four parts: initial feature extraction, coarse fusion, fine fusion, and rain streak reconstruction, which are combined to regress the residual rain image $I^{*}_{R}$. We produce the rain-free image $I_{Derain}$ by subtracting $I^{*}_{R}$ from the original rain image $I_{Rain}$. The goal is to make $I_{Derain}$ as close as possible to the rain free image $I_{Clean}$.}
\label{fig:MSPFN}
\end{figure*}

To address these limitations of the prior works, we explore the multi-scale representation from input image scales and deep neural network representations in a unified framework, and propose a multi-scale progressive fusion network (MSPFN) to exploit the correlated information of rain streaks across scales for single image deraining. Specifically, we first generate the Gaussian pyramid rain images using Gaussian kernels to down-sample the original rain image in sequence. A coarse-fusion module (CFM) (\textcolor{red}{\S 3.1}) is designed to capture the global texture information from these multi-scale rain images through recurrent calculation (Conv-LSTM), thus enabling the network to cooperatively represent the target rain streak using similar counterparts from global feature space. Meanwhile, the representation of the high-resolution pyramid layer is guided by previous outputs as well as all low-resolution pyramid layers. A fine-fusion module (FFM) (\textcolor{red}{\S 3.2}) is followed to further integrate these correlated information from different scales. By using the channel attention mechanism, the network not only discriminatively learns the scale-specific knowledge from all preceding pyramid layers, but also reduces the feature redundancy effectively. Moreover, multiple FFMs can be cascaded to form a progressive multi-scale fusion. Finally, a reconstruction module (RM) is appended to aggregate the coarse and fine rain information extracted respectively from CFM and FFM for learning the residual rain image, which is the approximation of real rain streak distribution. The overall framework is outlined in Fig.~\ref{fig:MSPFN}.
The main contributions of this paper are as follows:
\begin{itemize}
\item We uncover the correlations of rain streaks in an image and propose a novel multi-scale progressive fusion network (MSPFN) which collaboratively represents rain streaks from multiple scales via the pyramid representation.
\vspace{-5pt}
\item To better characterize rain streaks of different scales, we devise three basic modules, coarse-fusion module (CFM), fine-fusion module (FFM) and reconstruction module (RM), to effectively extract and integrate the multi-scale information. In these modules, the complementary information of similar patterns with rain streaks, both within the same scale or across different scales (pyramid layers), is progressively fused to characterize the rain streaks distribution in a collaborative/cooperative manner.
\vspace{-5pt}
\item Apart from achieving the state-of-the-art deraining performance in terms of the conventional quantitative measurements (\eg PSNR and SSIM), we build several synthetic rain datasets based on COCO~\cite{caesar2018coco} and BDD~\cite{yu2018bdd100k} datasets for joint image deraining, detection and segmentation tasks. To the best of our knowledge, we are the first to apply mainstream vision-oriented tasks (detection and segmentation) for comprehensively evaluating the deraining performance.
\end{itemize}
\section{Related Work}
In the last few years, substantial improvements~\cite{qian2018attentive, li2019single, chen2018robust, li2019heavy} have been observed on rain image restoration. In this work, we mainly focus on single image deraining because it is more challenging.
\subsection{Single Image Deraining}
Previous traditional methods for single image deraining~\cite{chen2013generalized, kang2011automatic} fail under the complex rain conditions and produce degraded image contents due to the limited linear-mapping transformation. Very recently, deep-learning based approaches~\cite{qian2018attentive, wang2019spatial, zhang2017convolutional} have emerged for rain streak removal and demonstrated impressive restoration performance. For example, Fu~\etal~\cite{fu2017clearing} introduce a three-layer convolutional neural network (CNN) to estimate and remove rain streaks from its rain-contaminated counterpart. To better represent rain streaks, Zhang~\etal~\cite{zhang2018density} take the rain density into account and present a multi-task CNN for joint rain density estimation and deraining. Later, Zhang~\etal~\cite{zhang2019image} further incorporate quantitative, visual and discriminative performance into the objective function, and propose a conditional generative adversarial network for rain streak removal. In order to alleviate the learning difficulty, recurrent frameworks~\cite{li2018recurrent, yang2019single, ren2019progressive} are designed to remove rain streaks in a stage-wise manner.
\subsection{Multi-scale Learning}
Rain streaks in the air show the apparent self-similarity, both within the same scale or across different scales, which makes it possible to exploit the correlated information across scales for rain streak representation. However, most existing deraining methods~\cite{li2017single, zhang2018density} ignore the underlying correlations of rain streaks across different scales. Only a few attempts~\cite{fu2019lightweight, zhengresidual} have been made to exploit the multi-scale knowledge. Fu~\etal~\cite{fu2019lightweight} decompose the restoration task into multiple subproblems and employ a set of parallel subnetworks to individually estimate the rain information in a specific pyramid scale space. However, it does not exploit and utilize the correlated information among these pyramid layers. Different from the parallel pyramid framework in~\cite{fu2019lightweight}, Zheng~\etal~\cite{zhengresidual} propose the cascaded pyramid network, which is similar to LapSRN~\cite{lai2017deep}, to iteratively remove rain streaks. However, only the high-level features are used to help the adjacent pyramid representation, which results in losing some useful hierarchical and scale features in a deep cascaded network. The significance of these features produced at different stages has been verified on image reconstruction tasks~\cite{tong2017image, zhang2018residual}.

Different from these methods~\cite{fu2019lightweight, zhengresidual}, in this work we introduce a novel framework MSPFN to achieve the collaborative representation of rain streaks across different scales, where the rich multi-scale rain information extracted from the Gaussian pyramid images is progressively aggregated along the pyramid layers and stages of the network. As a result, our predicted rain streak distribution is more accurate via the multi-scale collaborative representation.

\section{Proposed Method}
Fig.~\ref{fig:MSPFN} shows the overall pipeline of our proposed multi-scale progressive fusion network (MSPFN) for image deraining by excavating and exploiting the inherent correlations of rain streaks across different scales. We present the details of each building block and the loss function in the following.
\begin{figure}[!t]
\centering
\includegraphics[width=3.2in]{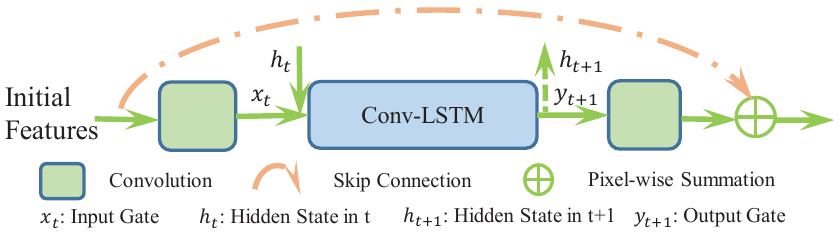}
\caption{Pipeline of the proposed residual recurrent units (RRU).}
\label{fig:RRU}
\end{figure}
\subsection{Multi-scale Coarse Fusion}
For a given rain image, our method first generates the Gaussian pyramid rain images using Gaussian kernels to down-sample the original rain image into different scales, \eg 1/2 and 1/4. The network takes as input the pyramid rain images and extracts the shallow features through multiple parallel initial convolution layers (see the first block of ``initial layer" in Fig.~\ref{fig:MSPFN}). Based on the initial features from each scale, the coarse-fusion module (CFM) then performs the deep extraction and fusion of multi-scale rain information through several parallel residual recurrent units (RRU), as shown in Fig.~\ref{fig:RRU}. The reasons for designing CFM are three folds: (a) To exploit the repetition of rain streaks under the same scale, we apply the recurrent calculation and residual learning to capture the global texture information, making it possible to cooperatively represent target rain streaks. More accurately, we introduce Conv-LSTM to model the information flow of context textures at spatial dimension with the recursive memory, where the contextual texture correlations are transformed into structured cyclic dependencies to capture the complementary or redundant rain information (\eg the solid arrows in Fig.~\ref{fig:example}). (b) The multi-scale structure provides an alternative solution to greatly increase the receptive filed to cover more contents while maintaining a shallow depth. (c) The high-resolution representations benefit from the outputs of previous stages as well as all low-resolution pyramid layers via iterative sampling and fusion.
\subsection{Multi-scale Fine Fusion}
\begin{figure}[!t]
\centering
\includegraphics[width=3.2in]{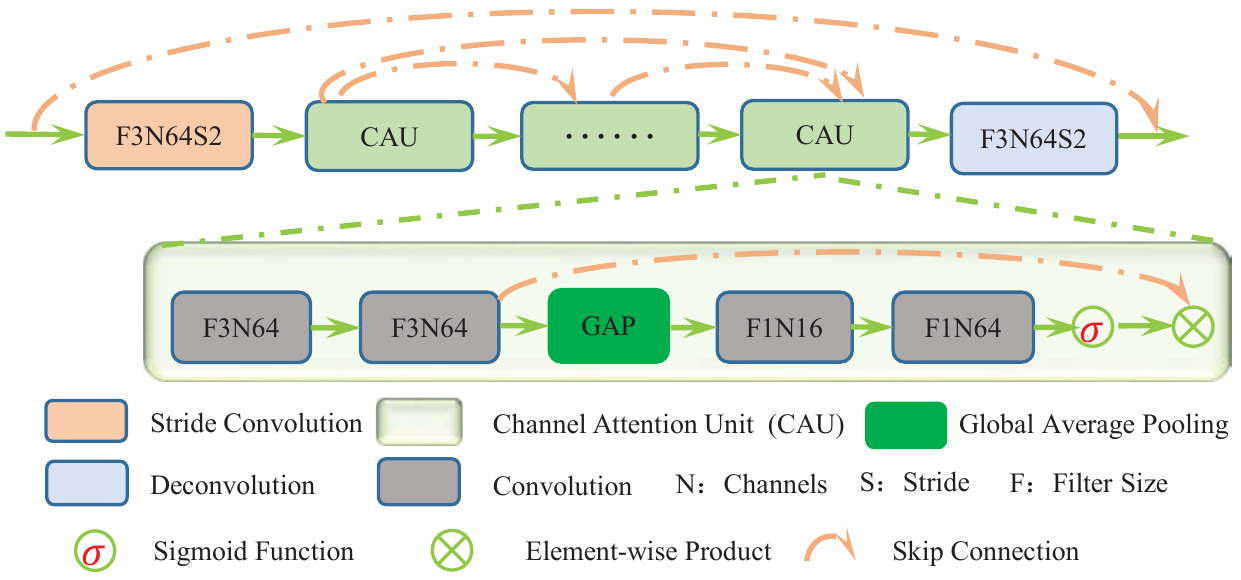}
\caption{Pipeline of our proposed U-shaped residual attention block (URAB). URAB is composed of several cascaded channel attention units (CAUs) to promote the fusion of the multi-scale rain information and reduce the feature redundancy by focusing on the most useful channels.}
\label{fig:FFM}
\end{figure}

The outputs of CFM go through the fine-fusion module (FFM) to refine the correlated information from different scales. As shown in Fig.~\ref{fig:MSPFN}, FFM enjoys the similar multi-scale structure with CFM for convenience. Unlike CFM, we introduce the channel attention unit (CAU) to enhance the discriminative learning ability of the network through focusing on the most informative scale-specific knowledge, making the cooperative representation more efficient. To alleviate the computation burden, we apply the strided convolution to reduce the spatial dimension of features, and finally utilize the deconvolution layer to increase the resolution to avoid losing resolution information, resulting in the U-shaped residual attention block (URAB). As depicted in Fig.~\ref{fig:FFM}, URAB is composed of several CAUs, along with the short skip connections to help the fine representation of multi-scale rain information. Moreover, long skip connections are used between cascaded FFMs to achieve progressive fusion of multi-scale rain information as well as to facilitate the effective backward propagation of the gradient.

\subsection{Rain Streak Reconstruction}
To learn the final residual rain image, we further integrate both low- and high-level multi-scale features respectively from CFM and FFM via a reconstruction module (RM), schematically depicted in Fig.~\ref{fig:MSPFN}. Specifically, the outputs from CFM are concatenated with the outputs from the last FFM, and then a convolution layer is used to learn the channel interdependence and rescale the feature values from the two modules. Similarly, the iterative sampling and fusion of rain information across different pyramid layers are implemented to estimate the residual rain image.
\subsection{Loss Function}
Mean squared error (MSE) is the commonly used loss to train the network~\cite{zhang2018density,yang2019scale}. However, it usually produces blurry and over-smoothed visual effect with the loss of high-frequency textures due to the squared penalty. In this work, we perform the successive approximation to the real rain streak distribution $I_{R}$ with the guidance of the Charbonnier penalty function~\cite{lai2017deep}, which is more tolerant of small errors and holds better convergence during training. The function is expressed as
\begin{equation}
\emph{L}_{con} = \sqrt{(I^*_{R}- I_{R})^2+\varepsilon^2}.
\label{eq:loss1}
\end{equation}
In Equation~(1), $I^*_{R}$ denotes the predicted residual rain image. The predicted rain-free image $I_{Derain}$ is generated by subtracting $I^*_{R}$ from its rain-contaminated counterpart $I_{Rain}$. The penalty coefficient $\varepsilon$ is empirically set to $10^ {-3}$.

In order to further improve the fidelity and authenticity of high-frequency details while removing rain streaks, we propose the additional edge loss to constrain the high-frequency components between the ground truth $I_{Clean}$ and the predicted rain-free image $I_{Derain}$. The edge loss is defined as
\begin{equation}
\emph{L}_{edge} = \sqrt{(Lap(I_{Clean})- Lap(I_{Derain}))^2+\varepsilon^2}.
\label{eq:loss2}
\end{equation}
In Equation~(2), $Lap(I_{Clean})$ and $Lap(I_{Derain})$ denote the edge maps respectively extracted from $I_{Clean}$ and $I_{Derain}$ via the Laplacian operator~\cite{kamgar1999optimally}. Then, the total loss function is given by
\begin{equation}
\emph{L} = L_{con} + \lambda \times L_{edge},
\label{eq:loss3}
\end{equation}
where the weight parameter $\lambda$ is empirically set to 0.05 to balance the loss terms.
\section{Experiments and Discussions}
We conduct extensive experiments on several synthetic and real-world rain image datasets~\cite{fu2017removing,zhang2019image,wang2019spatial} to evaluate the restoration performance of our proposed MSPFN as well as six state-of-the-art deraining methods. These representative methods include DerainNet~\cite{fu2017clearing}, RESCAN~\cite{li2018recurrent}, DIDMDN~\cite{zhang2018density}, UMRL~\cite{yasarla2019uncertainty}, SEMI~\cite{wei2019semi} and PreNet~\cite{ren2019progressive}. There is no unified training datasets for all competing methods in this paper, e.g. PreNet refers to JORDER~\cite{yang2017deep} and uses 1254 pairs for training. UMRL refers to~\cite{zhang2018density} and uses 12700 images for training. Therefore, directly taking the results from their papers is unfair and meaningless. To this end, we collect about $13700$ clean/rain image pairs from~\cite{zhang2019image,fu2017removing} for training our network as well as other competing methods for a fair comparison. In particular, these competing methods are retrained in the experiments with their publicly released codes and follow their original settings under the unified training dataset. Separately, the detailed descriptions of the used datasets are tabulated in Table~{\ref{table:datasets}}. In order to quantitatively evaluate the restoration quality, we adopt the commonly used evaluation metrics, such as Peak Signal to Noise Ratio (PSNR), Feature Similarity (FSIM)~\cite{zhang2011fsim}, and Structural Similarity (SSIM)~\cite{wang2004image}.
\subsection{Implementation Details}
\begin{table}
\begin{center}
\caption{Dataset description. A total of 13712 clean/rain image pairs are used for training. There are additional 4300 labeled reference samples as well as 200 real-world scenarios for testing.}
\label{table:datasets}
\scriptsize
\begin{tabular}{|l|c|c|c|}
\hline
Datasets & Training Samples & Testing Samples& \textbf{Name}\\
\hline\hline
Rain14000~\cite{fu2017removing} & 11200 & 2800 & \textbf{Test2800}\\
\hline
Rain1800~\cite{yang2017deep} & 1800 & 0 & \textbf{Rain1800}\\
\hline
Rain800~\cite{zhang2019image} & 700 & 100 & \textbf{Test100}\\
\hline
Rain100H~\cite{yang2017deep} & 0 & 100 & \textbf{Rain100H}\\
\hline
Rain100L~\cite{yang2017deep} & 0 & 100 & \textbf{Rain100L}\\
\hline
Rain1200~\cite{zhang2018density} & 0 & 1200 & \textbf{Test1200}\\
\hline
Rain12~\cite{li2016rain} & 12 & 0 & \textbf{Rain12}\\
\hline
Real200~\cite{wang2019spatial,wei2019semi} & 0 & 200 & \textbf{Real200}\\
\hline
RID/RIS~\cite{li2019single} & 0 & 2495/2348 & \textbf{RID/RIS}\\
\hline\hline
Total Count & 13712 & 9343 & - \\
\hline
\end{tabular}
\end{center}
\end{table}

In our baseline, the pyramid levels are set to 3, \ie the original scale, 1/2 scale and 1/4 scale. In CFM, the filter numbers of each recurrent Conv-LSTM are respectively set to 32, 64, and 128, corresponding to the gradually increasing resolution. The depths/numbers of FFM (\textbf{M}) and CAU (\textbf{N}) are set to 10 and 3, respectively. We use Adam optimizer with batch size of 8 for training on one NVIDIA Titan Xp GPU. The learning rate is initialized to $2\times10^{-4}$ and reduced by half at every 20000 steps till $1\times10^{-6}$. We train the network for 30 epochs with the above settings.
\begin{table}[t]
\begin{center}
\caption{Evaluation of the basic components in our baseline MSPFN on \textbf{Test100} dataset. We obtain the average inference time of deraining on images with size of \textbf{512$\times$ 384}.}
\label{table:MFFM}
\tiny
\begin{tabular}{|l|c|c|c|c|c|c|c|}
\hline
Models & Model1 & Model2 & Model3 & Model4 & Model5 & Model6 & MSPFN\\
\hline\hline
PSNR & 26.56 & 27.01 & 23.69 & 26.75 & 26.48 & 26.88 &  \textbf{27.29}\\
\hline\hline
SSIM & 0.861 & 0.864 & 0.831 & 0.863 & 0.862 & 0.865 & \textbf{0.869}\\
\hline\hline
FSIM & 0.921 & 0.923 & 0.905 & 0.923 & 0.921 & 0.923 &  \textbf{0.925}\\
\hline\hline
Ave. inf. time (s) & 0.192 & 0.224 & \textbf{0.113} & 0.238 & 0.141 & 0.180 &  0.308\\
\hline\hline
Par. (Millions) & 5.53 & 11.30 & \textbf{2.29} & 11.75 & 5.60 & 8.45 &  13.22\\
\hline
\end{tabular}
\end{center}
\end{table}

\subsection{Ablation Studies}
\begin{table*}[t]
\begin{center}
\caption{Evaluation of the depth of FFM (\textbf{M}), the number of CAU (\textbf{N}), as well as the model parameters on \textbf{Test100} dataset. MSPFN$_{MaNb}$ denotes the model with \textbf{M} $= a$ and \textbf{N} $= b$.}
\label{table:RDM}
\scriptsize
\begin{tabular}{|l|c|c|c|c|c|c|c|}
\hline
Models & MSPFN$_{M30N1}$ & MSPFN$_{M17N1}$ &  MSPFN$_{M17N2}$ &  MSPFN$_{M13N2}$ & MSPFN$_{M10N3}$&  MSPFN$_{M8N5}$ & MSPFN$_{M5N1}$\\
\hline\hline
PSNR & \textbf{27.91}  & 27.50  &27.63 & 27.42& 27.29 & 27.13 & 24.99\\
\hline\hline
SSIM & \textbf{0.879}  & 0.876 & 0.877& 0.874& 0.869  & 0.867 & 0.850\\
\hline\hline
SSIM & \textbf{0.929} & 0.928 &0.928 & 0.927& 0.925 & 0.924 & 0.916\\
\hline\hline
Par. (Millions) & 21.81 & 13.35 & 17.20 & 13.63& 13.22 & 14.56 & \textbf{1.65}\\
\hline
\end{tabular}
\end{center}
\end{table*}
\textbf{Validation on Basic Components.} Using our baseline model ($\textbf{M}= 10$, $\textbf{N} = 3$), we design six comparison models to analyze the effects of the proposed basic modules (CFM and FFM), multi-scale pyramid framework, and multi-scale progressive fusion scheme on deraining performance. Quantitative results on \textbf{Test100} dataset are listed in Table~{\ref{table:MFFM}}. From the results, our baseline MSPFN exhibits great superiority over its incomplete versions, including Model1 (single-scale framework with only the original input), Model2 (removing CFM from MSPFN), and Model3 (removing all FFMs from MSPFN), surpassing them by 0.73dB, 0.28dB, and 3.60dB (PSNR), respectively. Moreover, we construct Model4 by applying the fusion strategy in~\cite{fu2019lightweight} to verify the effectiveness of the proposed multi-scale progressive fusion scheme. It is evident that MSPFN gains a significant improvement over Model4 by 0.54dB with an acceptable complexity increase. Model5 ($\textbf{M}= 5$, $\textbf{N} = 1$) and Model6 ($\textbf{M}= 6$, $\textbf{N} = 3$) are the simplified variants of MSPFN with smaller depths. When compared with the single-scale framework (Model1), Model5 has the approximately equal amount of parameters but achieves faster inference speed with the multi-scale pyramid framework. Model6 has the similar computation complexity but more parameters as compared with Model1. The results show that Model5 achieves the comparable performance while it's a quarter more efficient. Model6 gains the better scores over Model1 by 0.32dB while keeping the similar computation complexity. We attribute these advantages to the effective cooperative representation of rain streaks among different pyramid layers and stages of the network.

\begin{figure*}[!t]
\centering
\vspace{-12pt}
\includegraphics[width=5.3in]{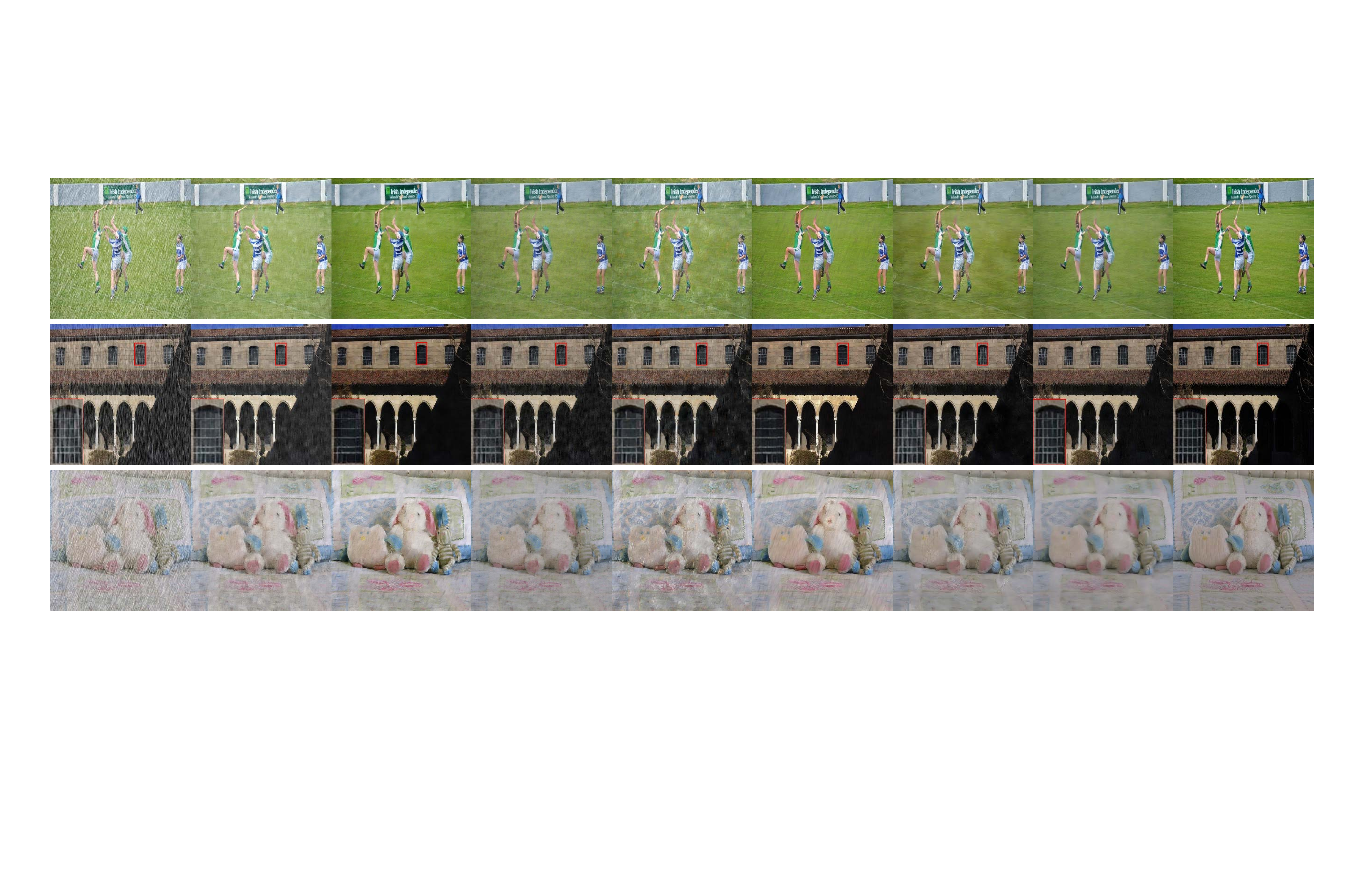}\\
\vspace{-4pt}
\includegraphics[width=5.3in]{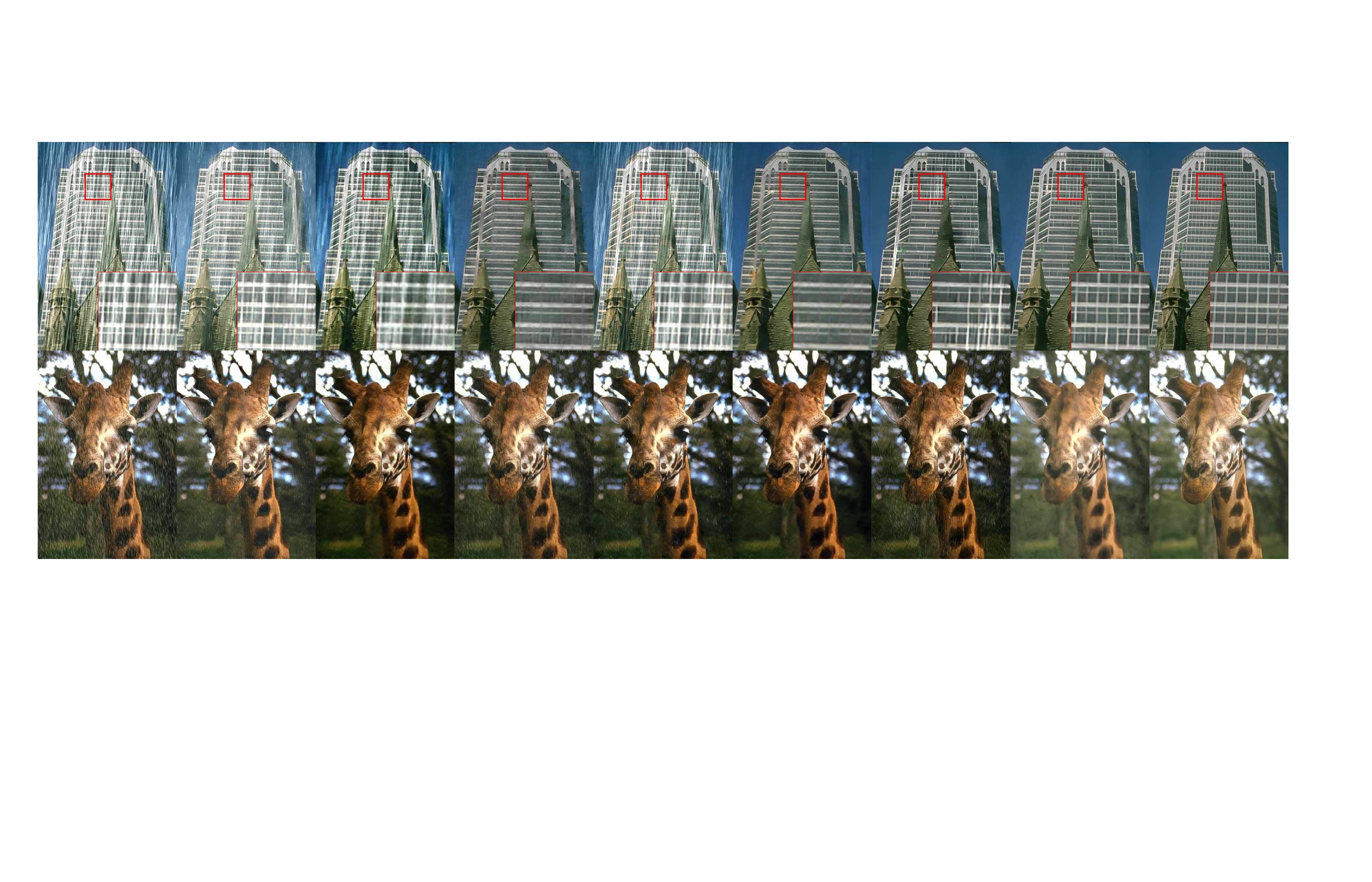}\\
\vspace{-4pt}
\includegraphics[width=5.3in]{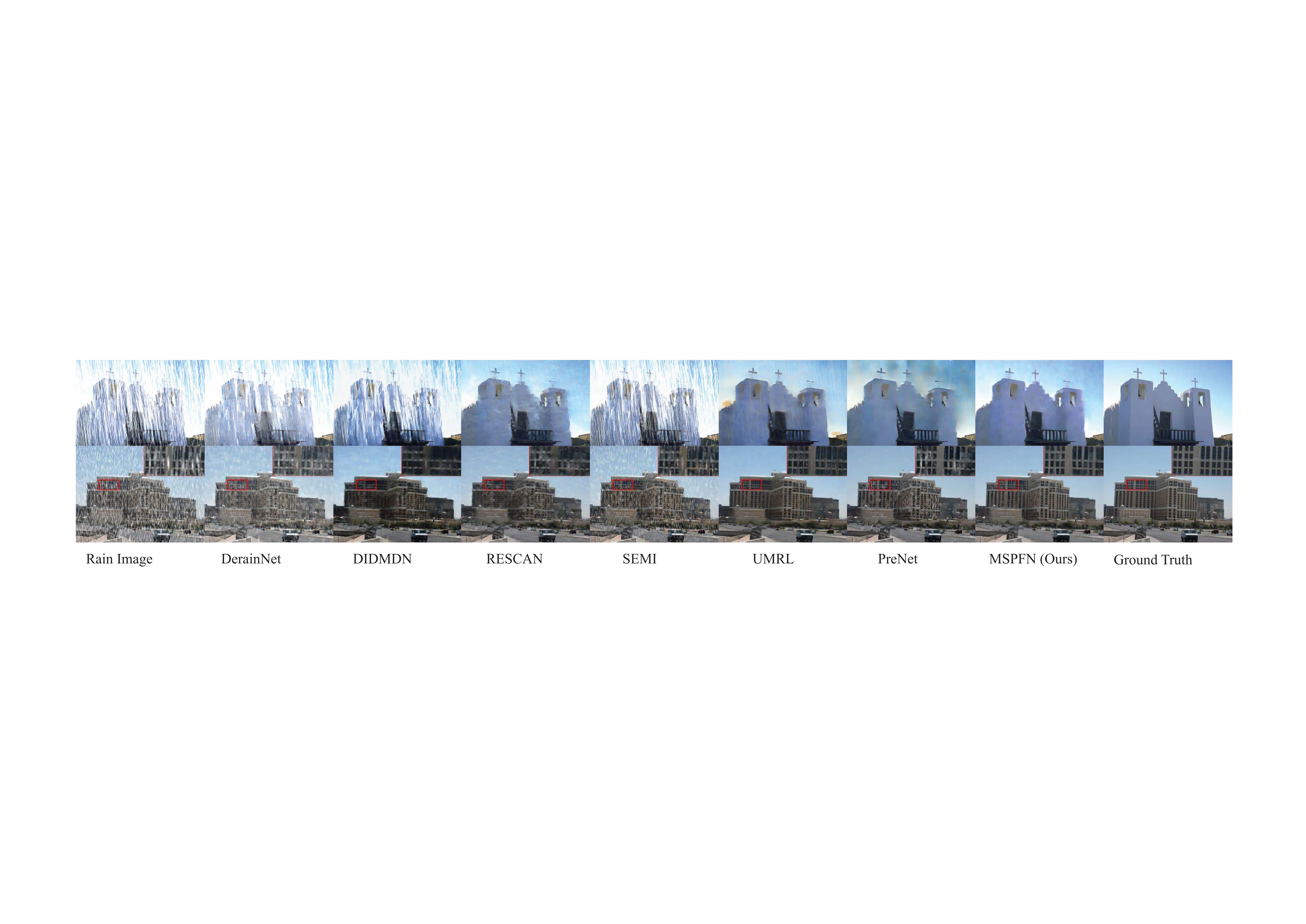}
\vspace{-2pt}
\caption{Restoration results on synthetic datasets, including \textbf{Rain100H}, \textbf{Rain100L}, \textbf{Test100}, and \textbf{Test1200}.}
\label{fig:synthetic}
\end{figure*}

\begin{table*}
\begin{center}
\caption{Comparison results of average PSNR, SSIM and FSIM on several widely used rain datasets, including \textbf{Rain100H}, \textbf{Rain100L}, \textbf{Test100}, \textbf{Test2800}, and \textbf{Test1200}. \textbf{MSPFN$_{w/o\ Eloss}$ denotes our model without the edge constraint in the loss function}.}
\label{table:synthetic}
\scriptsize
\begin{tabular}{|c|c|c|c|c|c|c|}
    \hline
    \multirow{2}*{Methods} & \textbf{Test100}&\textbf{Rain100H}&\textbf{Rain100L}&\textbf{Test2800}&\textbf{Test1200}&Average\\
     &PSNR/SSIM/FSIM& PSNR/SSIM/FSIM& PSNR/SSIM/FSIM& PSNR/SSIM/FSIM& PSNR/SSIM/FSIM& PSNR/SSIM/FSIM\\ \hline
    DerainNet~\cite{fu2017clearing} & 22.77/0.810/0.884& 14.92/0.592/0.755& 27.03/0.884/0.904& 24.31/0.861/0.930& 23.38/0.835/0.924& 22.48/0.796/0.879 \\
    \hline\hline
    RESCAN~\cite{li2018recurrent} & 25.00/0.835/0.909& 26.36/0.786/0.864& 29.80/0.881/0.919& 31.29/0.904/0.952& 30.51/0.882/0.944& 28.59/0.857/0.917 \\
    \hline\hline
    DIDMDN~\cite{zhang2018density} & 22.56/0.818/0.899& 17.35/0.524/0.726& 25.23/0.741/0.861& 28.13/0.867/0.943& 29.65/0.901/0.950& 24.58/0.770/0.876 \\
    \hline\hline
    UMRL~\cite{yasarla2019uncertainty} & 24.41/0.829/0.910& 26.01/0.832/0.876& 29.18/0.923/0.940& 29.97/0.905/0.955& 30.55/0.910/0.955& 28.02/0.880/0.927 \\
    \hline\hline
    SEMI~\cite{wei2019semi} & 22.35/0.788/0.887& 16.56/0.486/0.692& 25.03/0.842/0.893& 24.43/0.782/0.897& 26.05/0.822/0.917& 22.88/0.744/0.857 \\
    \hline\hline
    PreNet~\cite{ren2019progressive} & 24.81/0.851/0.916& 26.77/0.858/0.890& \textbf{32.44/0.950/0.956} & 31.75/0.916/0.956& 31.36/0.911/0.955& 29.42/0.897/0.934 \\
    \hline\hline
    MSPFN$_{w/o\ Eloss}$ (Ours) & 26.93/0.865/0.924& 28.33/0.842/0.883& 32.18/0.928/0.939& 32.70/0.928/0.964& 32.22/0.914/0.958& 30.51/0.895/0.934 \\
    \hline\hline
    MSPFN (Ours) & \textbf{27.50/0.876/0.928}& \textbf{28.66/0.860/0.890}& 32.40/0.933/0.943& \textbf{32.82/0.930/0.966}& \textbf{32.39/0.916/0.960}& \textbf{30.75/0.903/0.937} \\
    \hline
\end{tabular}
\end{center}
\end{table*}
\textbf{Parameter Analysis on $M$ and $N$.} We assess the influence of the depth of FFM (\textbf{M}) and the number of CAU (\textbf{N}) on deraining performance. Based on our baseline ($\textbf{M}= 10$, $\textbf{N} = 3$), we construct three comparison models, \ie MSPFN$_{M17N1}$, MSPFN$_{M13N2}$ and MSPFN$_{M8N5}$, while keeping approximately the same number of parameters. As shown in Table~{\ref{table:RDM}}, the performance declines with the reduction of \textbf{M}. This indicates the important role of FFM for exploiting the multi-scale rain information in a progressive fashion. When increasing the number of CAU (MSPFN$_{M17N2}$), it yields a slight improvement (0.13dB), but with additional $30\%$ of the parameters. We also add two models MSPFN$_{M30N1}$ and MSPFN$_{M5N1}$ for comparison. The former is designed to pursue a better deraining performance with more FFMs to enhance multi-scale fusion, while the latter is a lightweight model with smaller depth ($\textbf{M}= 5$, $\textbf{N} = 1$) and width (all filter channels $=$ 32). Meanwhile, the strided convolution and deconvolution are employed twice in our proposed U-shaped residual attention block (URAB) of MSPFN$_{M5N1}$ to further alleviate the computation burden. As we expected, MSPFN$_{M30N1}$ achieves the best scores for all the metrics. MSPFN$_{M5N1}$ still obtains the acceptable performance, although being a much lighter network. \textit{Considering the tradeoff between efficiency and deraining performance, we set $\textbf{M}$ and $\textbf{N}$ to 17 and 1 respectively in the following experiments.}

\subsection{Comparisons with State-of-the-arts}
\subsubsection{\textbf{Synthesized Data}}
We compare our MSPFN ($M=17, N=1$) with other six top-performing deraining methods~\cite{fu2017clearing, li2018recurrent, zhang2018density, yasarla2019uncertainty, wei2019semi, ren2019progressive} on five synthetic datasets. Quantitative results are shown in Table~{\ref{table:synthetic}}. One can see that MSPFN achieves remarkable improvements over these state-of-the-art methods. For example, MSPFN surpasses DerainNet~\cite{fu2017clearing} and DIDMDN~\cite{zhang2018density} by 9.01dB and 2.74dB, respectively, in terms of PSNR on Test1200 dataset. Visual results on different rain conditions (diverse rain streak orientations and magnitudes) are presented in Fig.~\ref{fig:synthetic}. MSPFN exhibits impressive restoration performance on all scenarios, generating results with rich and credible image textures while removing main rain streaks. For other comparison methods, they tend to blur the image contents, or still leave some visible rain streaks. For example, only our MSPFN restores the clear and credible image details in the ``Giraffe" image, while the competing methods fail to remove rain streaks and their results have obvious color distortion.
\begin{figure*}[!t]
\centering
\vspace{-8pt}
\includegraphics[width=5.8in]{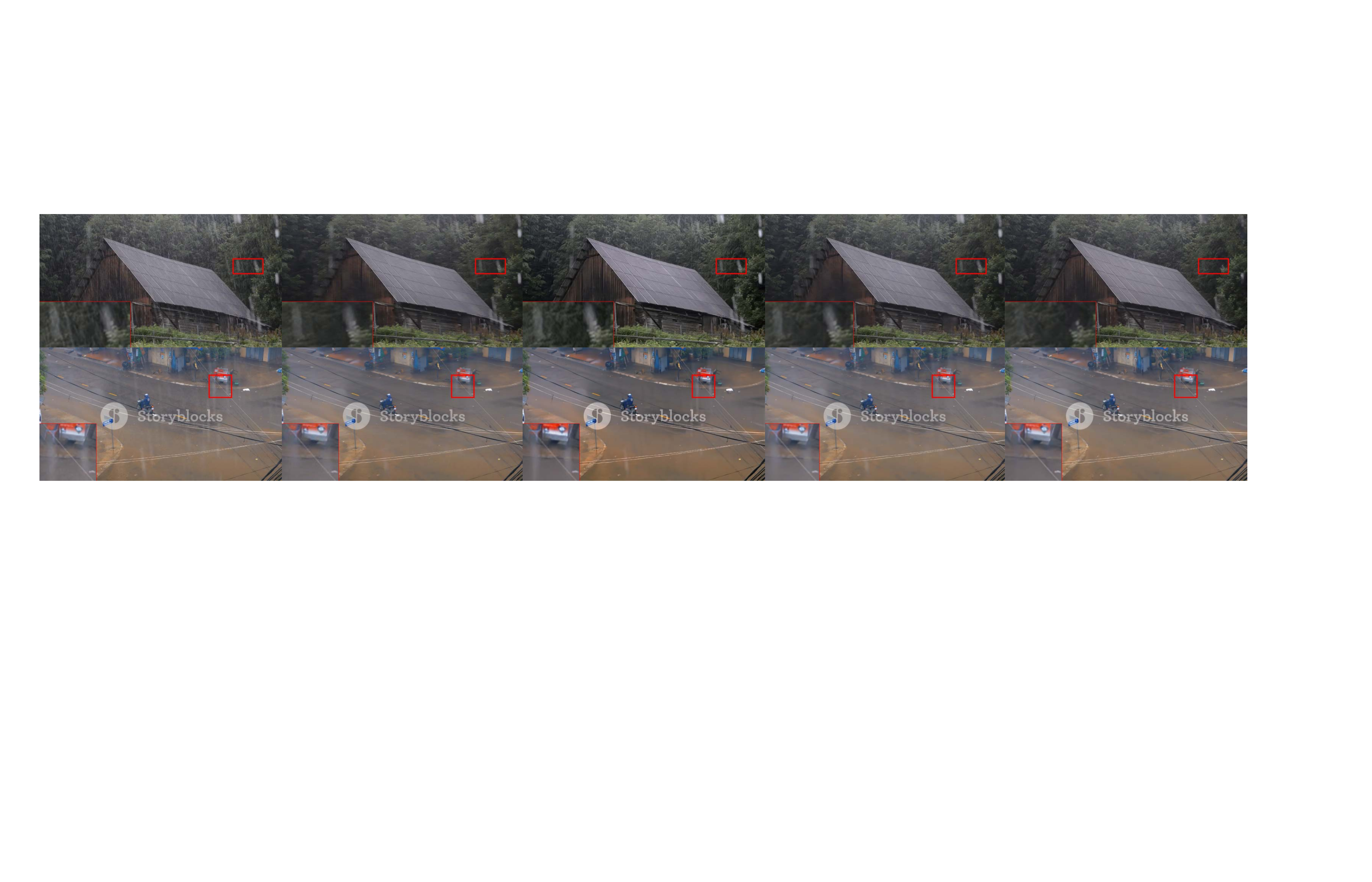}\\
\vspace{-3pt}
\includegraphics[width=5.8in]{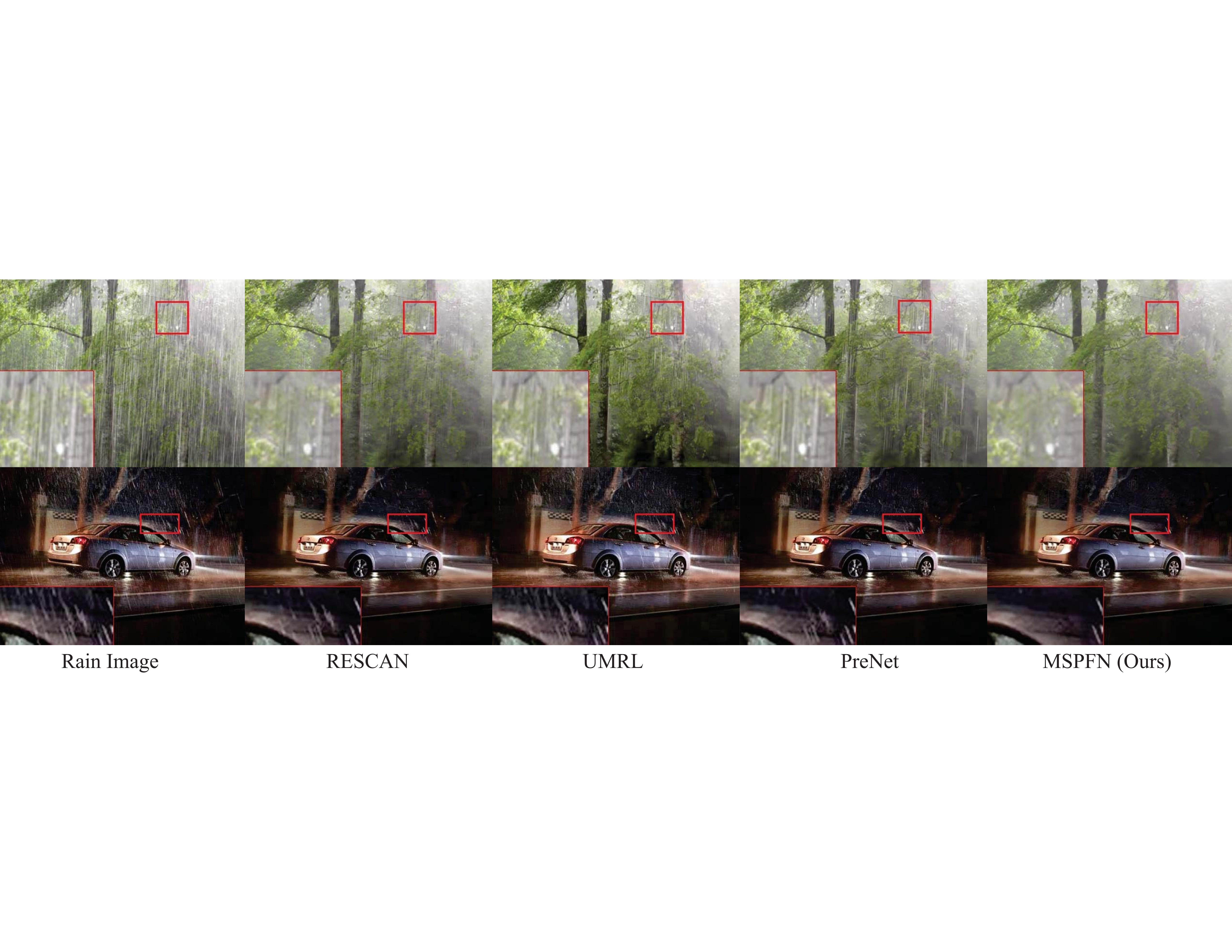}
\vspace{-2pt}
\caption{Comparison results on four real-world scenarios with RESCAN~\cite{li2018recurrent}, UMRL~\cite{yasarla2019uncertainty} and PreNet~\cite{ren2019progressive}.}
\label{fig:real}
\end{figure*}

\subsubsection{\textbf{Real-world Data}}
We conduct additional comparisons on three real-world datasets, including Real200~\cite{zhang2018density}, Rain in Driving (RID) and Rain in Surveillance (RIS) datasets~\cite{li2019single}, to further verify the generalization capability of MSPFN. RID and RIS cover 2495 and 2348 samples, collected from car-mounted cameras and networked traffic surveillance cameras in rainy days respectively. Moreover, we use another two quantitative indicators, Naturalness Image Quality Evaluator (NIQE)~\cite{6353522} and Spatial-Spectral Entropy-based Quality (SSEQ)~\cite{liu2014no}, to quantitatively evaluate the reference-free restoration performance. The smaller scores of SSEQ and NIQE indicate better perceptual quality and clearer contents. The results are listed in Table~{\ref{table:real}}. As expected, our proposed MSPFN has the best average scores on 200 real-world samples, outperforming the state-of-the-art deraining methods~\cite{li2018recurrent, yasarla2019uncertainty, ren2019progressive} by a large margin. Moreover, we show four representative deraining examples in Fig.~\ref{fig:real} for visual comparison. In the last image, obvious rain streaks are observed in the results of other deraining methods, but our MSPFN can well preserve more realistic and credible image details while effectively removing main rain streaks.
\begin{figure*}[!ht]
\centering
\includegraphics[width=6.0in]{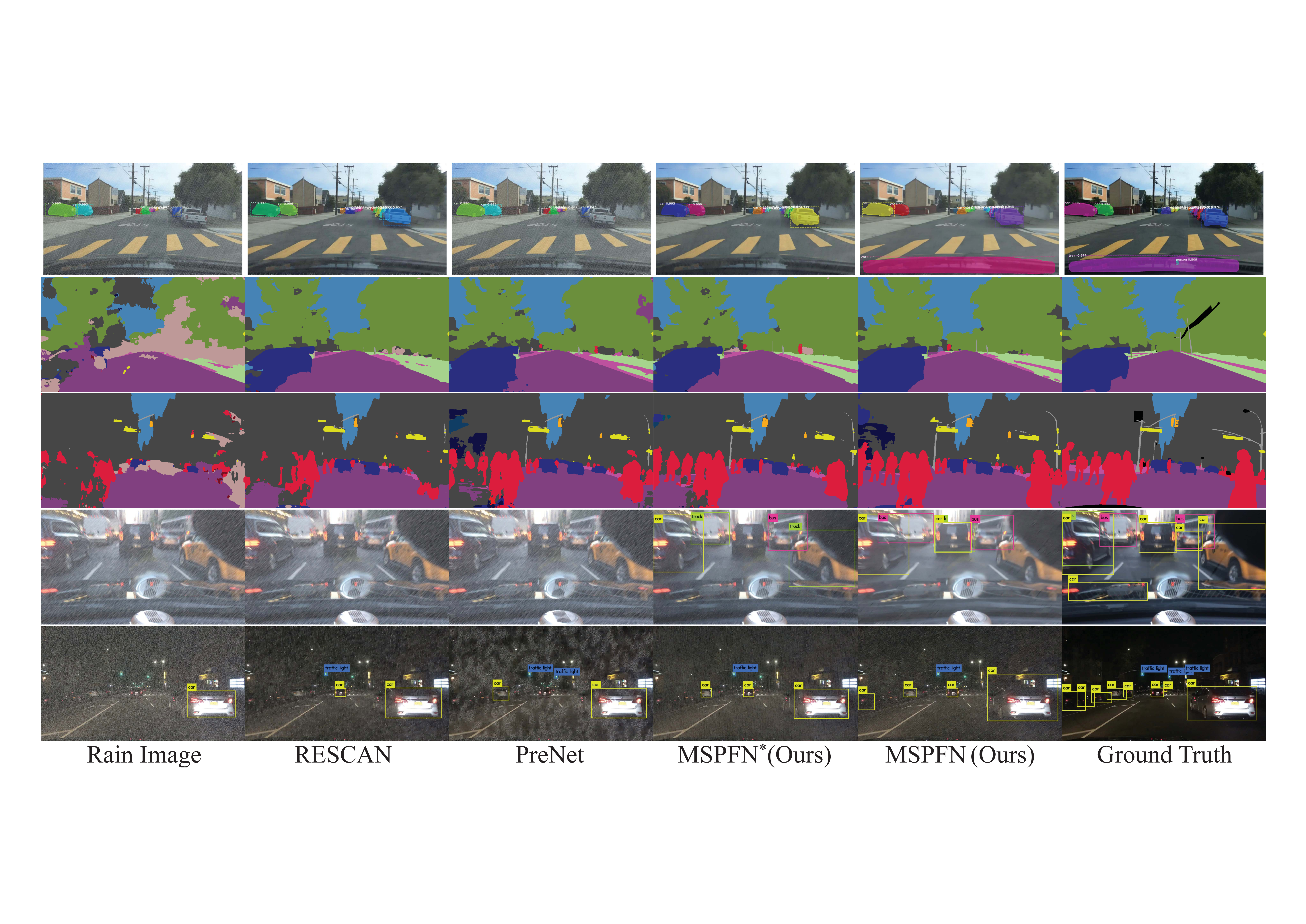}
\caption{Examples of joint deraining, object detection and segmentation. The first row denotes the instance segmentation results of Mask R-CNN~\cite{He2017ICCV} on BDD150 dataset. The second and third rows are the comparison results of semantic segmentation by RefineNet~\cite{lin2017refinenet} on BDD150 dataset. We use YOLOv3~\cite{redmon2018yolov3} for object detection on COCO350 dataset and the results are shown in the last two rows. MSPFN$^*$ denotes the lightweight model with lighter depth and width comparing to MSPFN.}
\label{fig:detection}
\end{figure*}
\begin{table}[!h]
\begin{center}
\caption{Comparison results of average NIQE/SSEQ on real-world datasets (\textbf{Real200,~RID,~and RIS}). The smaller scores indicate better perceptual quality.}
\label{table:real}
\scriptsize
\begin{tabular}{|l|c|c|c|c|}
\hline
Methods & RESCAN~\cite{li2018recurrent} & UMRL~\cite{yasarla2019uncertainty}& PreNet~\cite{ren2019progressive} & MSPFN (Ours)\\
\hline\hline
  Real200  & 4.724/30.47 & 4.675/29.38& 4.620/29.51& \textbf{4.459/29.26}\\
  \hline
  RID  & 6.641/40.62 & 6.757/41.04& 7.007/43.04& \textbf{6.518/40.47}\\
  \hline
  RIS  & 6.485/50.89 & \textbf{5.615/43.45}& 6.722/48.22& 6.135/43.47\\
\hline
\end{tabular}
\end{center}
\end{table}

\begin{table}[!t]
\begin{center}
\caption{Comparison results of joint image deraining, object detection, and semantic segmentation on COCO350, BDD350, and BDD150 datasets. MSPFN$^*$ denotes the lightweight model with lighter depth and width comparing to MSPFN.}
\label{table:detection}
\tiny 
\begin{tabular}{|l|c|c|c|c|c|}
\hline
Methods & Rain input& RESCAN~\cite{li2018recurrent}& PreNet~\cite{ren2019progressive} & MSPFN$^*$ (Ours)& MSPFN (Ours)\\
\hline\hline
\multicolumn{6}{|c|} {\textcolor[rgb]{0.50,0.00,1.00}{Deraining}; Dataset: \textbf{COCO350/BDD350}; Image Size: \textbf{640$\times$ 480}/\textbf{1280$\times$ 720}}\\
\cline{1-6}
\hline
  PSNR  & 14.79/14.13  &  17.04/16.71&  17.53/16.90 &  17.74/17.38 & \textbf{18.23/17.85}\\
  SSIM &  0.648/0.470   &  0.745/0.646  &  0.765/0.652 &  0.773/0.678 & \textbf{0.782/0.761}\\
   Ave.inf.time (s) &   --/-- &  0.55/1.53   &  0.22/0.76 &  \textbf{0.08/0.23} & 0.58/1.24\\
\hline\hline
\multicolumn{6}{|c|}{\textcolor[rgb]{1.00,0.00,0.00}{Object Detection}; Algorithm: \textbf{YOLOv3}~\cite{redmon2018yolov3}; Dataset: \textbf{COCO350/BDD350}; Threshold: 0.6}\\
\cline{1-6}
\hline
  Precision ($\%$) & 23.03/36.86  &  28.74/40.33  &  31.31/38.66 &  30.99/39.91 & \textbf{32.56/41.04}\\
  Recall ($\%$) &  29.60/42.80   &  35.61/47.79  &  37.92/48.59 &  37.99/49.74 & \textbf{39.31/50.40 }\\
  IoU ($\%$) & 55.50/59.85 &  59.81/61.98 &  60.75/61.08 &  61.06/61.90 & \textbf{61.69/62.42}\\
\hline\hline
\multicolumn{6}{|c|}{\textcolor[rgb]{0.50,0.00,1.00}{Deraining}; Dataset: \textbf{BDD150}; Image Size: \textbf{1280$\times$ 720}}\\
  \cline{1-6}
\hline
  PSNR & 18.00  &  20.96 &  21.52 &  21.73& \textbf{22.48}\\
  SSIM  &  0.722   &  0.859  &  0.886 &  0.887 & \textbf{0.904}\\
  Ave.inf.time (s) &   -- &  1.53   &  0.76 &  \textbf{0.23} & 1.24\\
\hline\hline
\multicolumn{6}{|c|}{\textcolor[rgb]{0.00,0.07,1.00}{Semantic Segmentation}; Algorithm: \textbf{RefineNet}~\cite{lin2017refinenet}; Dataset: \textbf{BDD150}}\\
  \cline{1-6}
\hline
  mPA ($\%$) &  33.29& 45.34  & 50.28& 50.25& \textbf{52.96}\\
  mIoU ($\%$) &  20.49& 31.52  & 33.42& 33.74& \textbf{35.90}\\
\hline
\end{tabular}
\end{center}
\end{table}
\vspace{-15pt}
\subsubsection{\textbf{Other Applications}}
\label{sec:op}
Image deraining under complex weather conditions can be considered as an effective enhancement of image content. It can potentially be incorporated into other high-level vision systems for applications such as object detection and segmentation. This motivates us to investigate the effect of restoration performance on the accuracy of object detection and segmentation based on some popular algorithms, \eg YOLOv3~\cite{redmon2018yolov3}, Mask R-CNN~\cite{He2017ICCV}, and RefineNet~\cite{lin2017refinenet}. To this end, we randomly select a total of 850 samples from COCO~\cite{caesar2018coco} and BDD~\cite{yu2018bdd100k} datasets to create three new synthetic rain datasets COCO350 (for detection), BDD350 (for detection), and BDD150 (for segmentation) through Photoshop. These rain images are of diverse streak orientations and magnitudes, and at the same time have complex imaging conditions such as night scenes. By using our proposed deraining algorithm MSPFN as well as other top-performing deraining methods~\cite{li2018recurrent,ren2019progressive}, the restoration procedures are directly implemented on these three datasets to produce the rain-free images. And then we apply the public available pre-trained models of YOLOv3 (for detection), Mask R-CNN (for instance segmentation), and RefineNet (for semantic segmentation) to perform the the downstream tasks. Qualitative results, including the deraining performance as well as the precision of the subsequent detection and segmentation tasks, are tabulated in Table~{\ref{table:detection}}. In addition, visual comparisons are shown in Fig.~\ref{fig:detection}.

It is obvious that rain streaks can greatly degrade the detection accuracy and segmentation precision, night scenarios in particular, \ie by missing targets and producing low detection or segmentation confidence (mean pixel accuracy (mPA) and  mean Intersection of Union (mIoU)). In addition, the detection precision of the produced rain-free images by MSPFN shows a notable improvement over that of original rain inputs by nearly 10\%, and MSPFN achieves the best results of $52.96\%$ mPA as well as $35.90\%$ mIoU for semantic segmentation task on BDD150. When compared with other top-performing deraining models, the rain-free images generated by MSPFN show more credible contents with more details, which effectively promote the detection and segmentation performance. Moreover, we also evaluate our lightweight deraining model MSPFN$^*$ with lighter depth ($\textbf{M}= 5$, $\textbf{N} = 1$) and width (with all filter channels of 32) since computation efficiency is crucial for mobile devices and applications require real-time throughput such as autonomous driving. MSPFN$^*$ still achieves competitive performance compared with other models~\cite{li2018recurrent,ren2019progressive} while it's a half more efficient in terms of inference time.

\section{Conclusion}
In this paper, we propose a novel multi-scale progressive fusion network (MSPFN) to exploit the multi-scale rain information to cooperatively represent rain streaks based on the pyramid framework. To achieve this goal, we design several basic modules (CFM, FFM and RM) along with our proposed multi-scale progressive fusion mechanism to explore the inherent correlations of the similar rain patterns among multi-scale rain streaks. Consequently, our predicted rain streak distribution is potentially more correct due to the collaborative representation of rain streaks across different scales. Experimental results on several synthetic deraining datasets and real-world scenarios, as well as several downstream vision tasks (\ie object detection and segmentation) have shown great superiority of our proposed MSPFN algorithm over other top-performing methods.
\section{Acknowledgement}
This work is supported by National Key R\&D Project (2016YFE0202300) and National Natural Science Foundation of China (U1903214, 61671332, U1736206, 41771452, 41771454, 61971165), and Hubei Province Technological Innovation Major Project (2019AAA049, 2018CFA024).
{\small
\bibliographystyle{ieee_fullname}
\bibliography{egbib}
}

\end{document}